\documentclass{article}

\usepackage{arxiv}

\usepackage[utf8]{inputenc} 
\usepackage[T1]{fontenc}    
\usepackage{hyperref}       
\usepackage{url}            
\usepackage{booktabs}       
\usepackage{amsfonts}       
\usepackage{nicefrac}       
\usepackage{microtype}      
\usepackage{lipsum}
\usepackage{times}
\usepackage{epsfig}
\usepackage{graphicx}
\usepackage{amsmath}
\usepackage{amssymb}
\usepackage{rotating}
\usepackage{multirow}
\usepackage{subfig}
\usepackage[norule,symbol,perpage]{footmisc}
\title{LivDet in Action - Fingerprint Liveness Detection Competition 2019}

\author{Giulia Orr\`{u}, Roberto Casula, Pierluigi Tuveri, Carlotta Bazzoni, Giovanna Dessalvi,\\Marco Micheletto,Luca Ghiani, and Gian Luca Marcialis\\
University of Cagliari - Department of Electrical and Electronic Engineering - Italy\\
{\tt\small \{ giulia.orru, pierluigi.tuveri, luca.ghiani, marcialis\}@diee.unica.it,}\\
}

\begin{document}
\maketitle

\begin{abstract}
The International Fingerprint liveness Detection Competition (LivDet) is an open and well-acknowledged meeting point of academies and private companies that deal with the problem of distinguishing images coming from reproductions of fingerprints made of artificial materials and images relative to real fingerprints. In this edition of LivDet we invited the competitors to propose integrated algorithms with matching systems. The goal was to investigate at which extent this integration impact on the whole performance. Twelve algorithms were submitted to the competition, eight of which worked on integrated systems.
\end{abstract}
\section{Introduction}

This paper reports the main relevant results obtained in the sixth edition of the International Fingerprint Liveness Detection Competition (http://livdet.diee.unica.it).

Born in 2009 with the joint efforts of the University of Cagliari and the Clarkson University, the number of participants and data publicly provided strongly grew up over the previous editions in 2011, 2013, 2015 and 2017 (from this edition Algorithms and Systems parts were separately leaded by University of Cagliari and the Clarkson University, respectively). 
In general, a Fingerprint Presentation Attack Detector (FPAD) or Fingerprint Liveness Detector (FLD) is a machine learning-based system able to prevent direct attacks to the sensor, by discriminating real and fake fingerprints, that is, fingerprint derived from artificial replica of alive fingers, also named artefacts.

Although these detectors may be designed as independent and stand-alone with respect to the problem of recognition or verification of the personal identity, recent works have pointed out the need to investigate their performance when integrated into real verification systems \cite{MarcelInAction}. Moreover, the need of having a ``liveness detection'' or ``anti-fake'' modality is important for companies interested in placing their products on the market of the computer security. 

The previous editions of the LivDet competition were limited to evaluating the performance of a fingerprint liveness detector operating alone. In real applications, the FPAD system works together with a recognition system in order to check whether the input comes from a live user or is a fake sample.
LivDet 2019 - LivDet in Action\footnote{Title inspired by the paper: ``Anti-spoofing in Action : Joint Operation with a Verification System'' \cite{MarcelInAction}.} invited the competitors to submit a complete algorithm able not only to output the liveness probability of the image given the extracted set of features (the so-called ``liveness score'') but also an integrated match score which includes the probability above (``integrated score''), on which basis the final user acceptance/rejection decision is taken. For this reason, this last edition was devoted to the fingerprint presentation attacks (liveness) detection ``in action''.

This paper is organized as follows. Section 2 makes an overview of fingerprint presentation attack detection with reference to the 2019 LivDet edition. Participants, datasets and the experimental protocol are shown in Section 3. Section 4 discusses the competition results. Conclusions are drawn in Section 5.

\section{Fingerprint Presentation Attack Detection (FPAD)}
   
\label{sec:Overview}
Software-based FPAD solutions relies on the extraction of anatomical, physiological or texture-based features from fingerprint images. The first work focused on these systems is from 1998 and it was written by D. Willis and M. Lee \cite{Scanners}. They discovered that four  biometric fingerprint devices out of six were ``vulnerable''  to spoofing attacks. Later, the issue was more detailed by Van der Putte \cite{vanderPutte2000} and Matsumoto \cite{MatsumotoMYH02}, which extended the investigation to other fingerprint scanners.
In the following years, in order to analyse and solve the spoofing problem, various works were published: in particular a hardware-based solution and the first software-based solution were proposed in the early 2000s \cite{kalloPatent,Schuckers2002SpoofingAA}.

Over the years, liveness detection methods have been refined and recently CNN-based methods have been proposed \cite{7390065,7029061,8296254}. 

Although the FPAD systems performance has achieved impressive results \cite{LivDet2017, IVC}, it is not yet clear the implication of their non-zero error when integrated into fingerprint verification/recognition applications \cite{MarcelInAction}.

For this reason, in this LivDet edition, besides the traditional treatment of the standard $\{Live, Fake\}$ classification problem, we investigated at which extent the integration of the liveness detector can impact on the whole performance of the fingerprint verification system.

\section{Participants, data sets and protocols}
\label{sec:Protocol}

\subsection{Participants}

The participants of LivDet are academies and companies. After the registration phase, each competitor must sign a license agreement detailing the proper usage of data released for the competition.
In this edition, it was possible to participate both with an integrated system and with a standard liveness detection system.
Table \ref{tab:partecipantsLivDet} shows the participants, the correspondent algorithms names adopted in this paper and the type of system presented.

\begin{table*}[h!]
\centering
\begin{tabular}{|c|c|c|}
\hline
\textbf{Participants} & \textbf{Algorithm names} & \textbf{Type}\\ \hline
Anonymous & LivDet19\_CNN & integrated\\ \hline
University of Naples Federico II & unina & integrated\\ \hline
CENATAV & PADUnkFv & liveness\\ \hline
Michigan State University & Fingerprint Spoof Buster [FSB]& liveness\\ \hline
Hangzhou Jinglianwen Technology Co.,Ltd. & JLW\_LivDet & liveness\\ \hline
Hangzhou Jinglianwen Technology Co.,Ltd. & JLWa & integrated\\ \hline
Hangzhou Jinglianwen Technology Co.,Ltd. & JLWs & integrated\\ \hline
Sichuan University & BRLFptLivDet & liveness\\ \hline
Inha University & halekim & integrated\\ \hline
Chosun University / Suprema ID Inc. & JungCNN & integrated\\ \hline
Zhejiang University of Technology & ZJUT\_Det\_ATHR & integrated\\ \hline
Zhejiang University of Technology & ZJUT\_Det\_S & integrated\\ \hline
\end{tabular}

\caption{ Name of the participants and the submitted algorithms.}
\label{tab:partecipantsLivDet}
\end{table*}

\subsection{Data Sets}

\begin{table*}[!t]
\centering
\begin{tabular}[t]{ | l | l | c | c | c | c |}
\hline
\textbf{Scanner} & \textbf{Model} & \textbf{Resolution [dpi]} & \textbf{Image Size [px]} & \textbf{Format} &\textbf{Type} \\ \hline
Green Bit & DactyScan84C & 500 & 500x500 & BMP & Optical \\ \hline
Orcanthus & Certis2 Image & 500 & 300x$n$  & PNG & Thermal swipe \\ \hline
Digital Persona & U.are.U 5160 & 500 &252x324 & PNG & Optical \\ \hline
\end{tabular}
\caption{Device characteristics for LivDet2019 datasets.}
\label{tab:sensors}
\end{table*}

\begin{figure}
\centering
  \includegraphics[height=10cm]{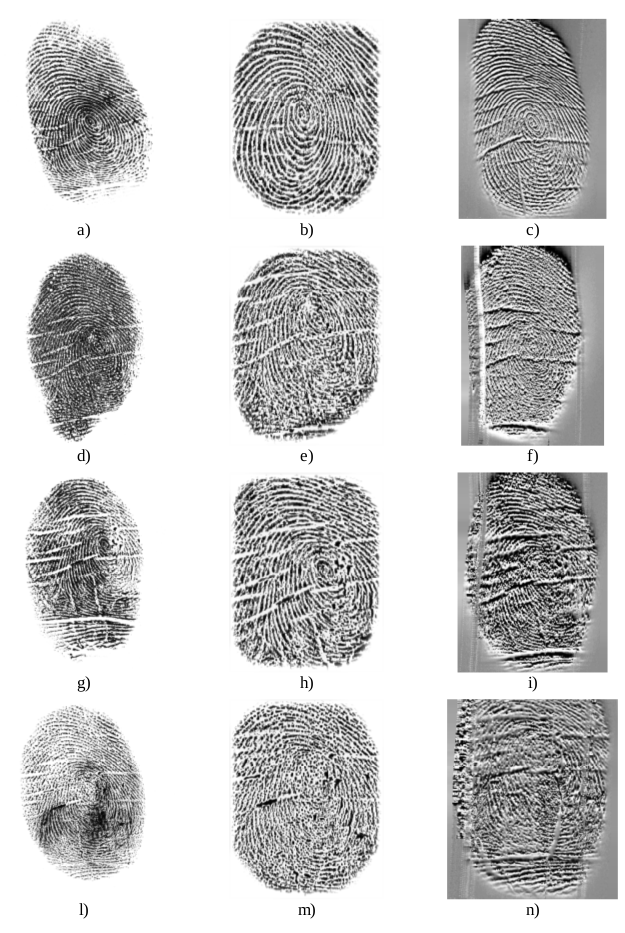}
  \caption{Examples of samples LivDet fingerprints dataset. In the first row live fingerprint for a) Green Bit b) Digital Persona and d) Orcanthus scanners. In the second row there is  Mix 2 spoof material for the e) Green Bit, f) Digital Persona and g) Orcanthus sensors. In the third row there are Liquid Ecoflex fake samples for the h) Green Bit, i) Digital Persona and f) Orcanthus devices. In the last row the samples of Mix 1 spoof material, with the same order of the other rows. We can appreciate in the third column the decay of the scanner. As matter of fact the Orcanthus is a thermal scanner and  the transducer has been progressively ruined due to by the excessive use thus there is a vertical band in the acquired image. }
  \label{fig:CampioniImmagini}
\end{figure}

\begin{table*}[t]
\centering
\resizebox{\textwidth}{!}{%
\begin{tabular}{|c||c|c|c|c|c|c||c|c|c|c||c|c|c|c|}
\hline
                 & \multicolumn{6}{c||}{\textbf{Train}}      & \multicolumn{4}{c||}{\textbf{Test}}\\ \hline
\textbf{Dataset} & Live & Wood Glue & Ecoflex & Body Double & Latex & Gelatine & Live & Mix 1 & Mix 2 & Liquid Ecoflex\\ \hline
Green Bit        &1000&400&400&400& - &-&1020&408&408&408\\ \hline
Orcanthus        &1000&400&400&400&-&-&990&384&308&396\\ \hline
Digital Persona  &1000&250&250&-&250&250&1019&408&408&408\\ \hline
\end{tabular}%
}
\caption{Number of samples for each scanner and each part of the dataset.}
\label{tab:datasetComposition}
\end{table*}
The training set for LivDet 2019 was composed by sub-set of the previous LivDet editions: the Orcanthus Certis2 Image and Green Bit DactyScan84C images from LivDet 2017 train set and the Digital Persona U.are.U 5160 images from LivDet 2015 train set. The scanner characteristics are reported in Table \ref{tab:sensors}.
It is important to note that the dimensions of the images acquired with the three sensors are very different from each other: this allows us to evaluate the performance of the algorithms on the basis of the acquisition surface and of the output images shape.
Indeed, as can be seen from the Figure \ref{fig:CampioniImmagini}, the images relating to the Digital Persona sensor (column 2) contain only a portion of the fingerprint while in the other two sensors the fingerprint is present in its entirety.

The LivDet 2019 fake images were collected using the cooperative method. Live images came from multiple acquisitions of at least six fingers of different subjects. The fakes creation methods can be classified as cooperative or non cooperative regarding the way the molds are collected. As a matter of fact, in the cooperative methods the mold is created through user collaboration, while in the non cooperative methods is made using a latent fingerprint.

Each dataset consists of two parts, the first is the training set, and the second one is the test set. The training set is released to participants in order to set the classification parameters of their algorithms. We use the test set once the participants uploaded their algorithms, as executable files.

The fake fingeprints inserted in the test set were created using materials different from those used in the train set. This consists of 6400 images, whilst the test set contains 6565 images. The number of samples for each scanner is shown in Table \ref{tab:datasetComposition}. As in the last edition, the fake samples of the training set and those of the test set are built by different members of the LivDet staff. In this way, we simulated a real scenario, where the ability of the attacker can be different from that of the laboratory staff. In this edition we created two new spoof material mix, leading to a never-seen-before material in the presentation attack detection.

\subsection{Algorithms Submission}

\begin{figure*}[htb]
\centering
\includegraphics[width=10cm]{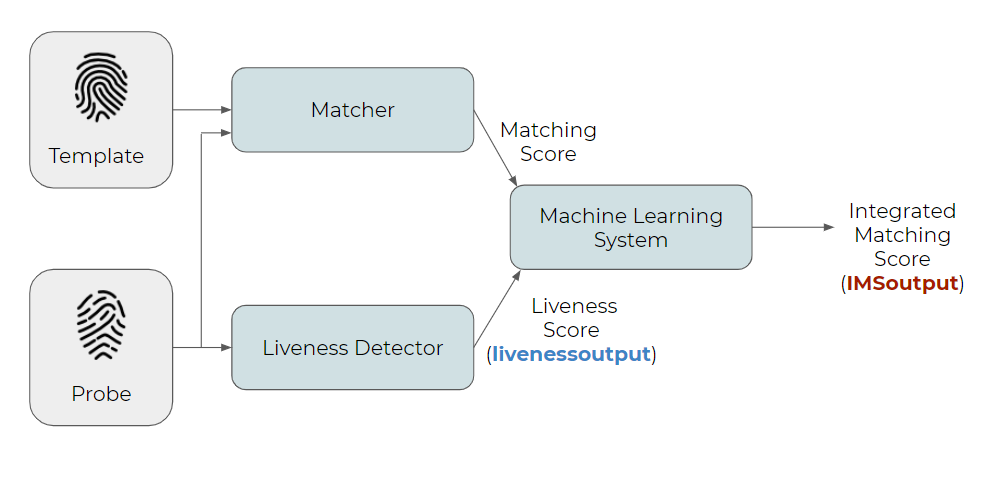}
\caption{\label{funzexample} Block diagram of a possible Integrated Verification System.}
\end{figure*}

The algorithms submission process for LivDet 2019 is different from that of previous editions. Each submitted algorithm is a console application with the following list of parameters:

\textit{[nameOfAlgorithm] [ndataset] [templateimagesfile] [probeimagesfile][livenessoutputfile] [IMSoutputfile]}
\\The parameter \textit{[ndataset]} is the identification number of the dataset to analyse, \textit{ [templateimagesfile]} is the text file name with the list of absolute paths of each template image registered in the system, while  \textit{ [probeimagesfile]} is the text file with the list of absolute paths of each image to analyse. The last two parameters are the path of the output files where the algorithm saves the result regarding every probe image. In the \textit{[livenessoutputfile]} the liveness output of each processed image is saved. This output is the degree of ``liveness'' normalized in the range 0 and 100 (100 is the maximum degree of liveness, 0 means that the image is fake). Scores [0, 50) classify fingerprint image as ``fake" while scores [50,100] classify fingerprint image as ``live"(in Fig. \ref{funzexample} in blue ``livenessoutput"). The \textit{[IMSoutputfile]} reports for each probe the combined probability of being a live fingerprint and of belonging to the declared identity, normalized in the range 0 and 100 (in Fig. \ref{funzexample} in red ``IMSoutput").  An IMSoutput between [0, 50) classify fingerprint image as ``fake" or belonging to an attacker while a value between [50,100] classify fingerprint image as ``live" and belonging to the declared user.

The selected classification threshold in order to measure the performance is 50. In both outputs if the algorithm has not been able to process the image, the corresponding value will be -1000.
For the submission of the liveness algorithms, the last edition protocol was followed \cite{LivDet2017}.
The parameters adopted for the performance evaluation are the following:
\begin{itemize}
    \item IMG\_accuracy: Rate of correctly classified genuine live fingerprints.
    \item IMI\_accuracy: Rate of correctly classified impostor live or genuine fake fingerprints.
    \item Ferrlive: Rate of misclassified live fingerprints.
    \item Ferrfake: Rate of misclassified fake fingerprints.
\end{itemize}
Ferrlive (corresponding to BPCER according to ISO/IEC 30107-1:2016 \cite{iso}) and Ferrfake (APCER \cite{iso}) are calculated from the livenessoutput, while IMG\_accuracy and IMI\_accuracy are calculated from IMSoutput (Integrated Match Score output).

The IMG score derived from the comparison of a live fingerprint with all the other acquisitions of the same fingerprint (same person, same finger). We decided to use all the possible combinations, because usually a fingerprint matcher is not symmetrical. In other words,  $s (a.b)\neq s (b, a)$, where $a$ and $b$ are any two fingerprints, and $s$ is any fingerprint matcher. 

The IMI score was calculated considering two types of comparisons: the first is the comparison between a live fingerprint and the corresponding fake fingerprint (same finger, same person), the second is the comparison between two live fingerprints of different people. The second class, due to reasons of excessive computational time, has been drastically reduced. 
The number of comparisons
made is shown in Table \ref{tab:comparison}.
Being the number of comparisons for calculating IMI and IMG different, the total IM accuracy was calculated as the weighted average of the latter. The comparisons number  with the Orcanthus scanner are smaller than with Green Bit and Digital Persona because of the least acquired images number, see Tab. \ref{tab:datasetComposition}.

All the comparisons were made with the templates belonging to live fingerprints, while the probes could be both live or fake in order to simulate a real scenario. It is worth mentioning that, although a fingerprint presentation attack using templates belonging to a fake sample cannot be excluded, we have not taken into consideration this particular case.

\begin{table}[]
\centering
\resizebox{0.49\textwidth}{!}{%

\begin{tabular}{|l|c|c|c|}
\hline
\multicolumn{4}{|c|}{\textbf{Number of IM scores per class (Genuine/Impostor)}}                                                          \\ \hline
\textbf{Scanner}      & \textbf{Green Bit}                 & \textbf{Digital Persona}                 & \textbf{Orcanthus}                 \\ \hline
\textbf{IMG scores}   & 4080                       & 4072                       & 3960                       \\ \hline
\textbf{IMI scores}   & 12852                      & 12836                      & 11776                      \\ \hline
\textbf{Total scores} & 16932                      & 16908                      & 15736 \\ \hline
\end{tabular}
 }
\caption{Number of integrated match scores (IM) for the accuracy of integrated systems. IMG scores are among live images of the same user. IMI scores simulate two types of attacks: attacks by an impostor (zero-effort), attacks by a fake fingerprint (spoofing, by a presentation attack instrument - PAI).}
\label{tab:comparison}
\end{table}

\section{Discussions on the reported results}

\label{sec:Result}

The liveness accuracy of each algorithm on the LivDet test sets, including the overall average accuracy, are summarized in Table \ref{tab:overalaccuracy}. Error rates on live and fake images are also reported. The performance of almost all algorithms on Green Bit and Orchantus datasets are over 90\%. In particular many algorithms reach 99\% of accuracy on the Green Bit dataset.
The images acquired with Digital Persona instead have been very difficult to classify and this has reduced the accuracy of many algorithms. In particular, the unina algorithm confuses more than 80\% of fake images with live, but the performance is below average for the other algorithms too.

It may be seen, in the Orchantus case, that generally the accuracy on the live fingeprints is worse than that on fake fingers. This strongly impacts when systems are integrated with matchers. By considering the unbalanced performances in terms of Ferrlive and Ferrfake, it may be hypothesized that the liveness threshold could be relaxed to obtain a better performance on genuine users, at the expense of misclassifying more fake fingerprints.

\begin{table*}[h]

\centering
\resizebox{\textwidth}{!}{%
\begin{tabular}{|l||c|c|c||c|c|c||c|c|c||r|}
\hline
\multicolumn{1}{|c||}{\multirow{3}{*}{\textbf{Algorithm}}} & \multicolumn{3}{c||}{\textbf{Green Bit}}   & \multicolumn{3}{c||}{\textbf{Digital Persona}}     & \multicolumn{3}{c||}{\textbf{Orchantus}}  & \multicolumn{1}{c|}{\multirow{3}{*}{\textbf{\begin{tabular}[c]{@{}c@{}}Overall\\ \\ {[}\%{]}\end{tabular}}}} \\ \cline{2-10}
\multicolumn{1}{|c||}{}    & \multicolumn{1}{c|}{\multirow{2}{*}{\textit{\begin{tabular}[c]{@{}c@{}}Ferrlive\\ {[}\%{]}\end{tabular}}}} & \multicolumn{1}{c|}{\multirow{2}{*}{\textit{\begin{tabular}[c]{@{}c@{}}Ferrfake\\ {[}\%{]}\end{tabular}}}} & \multirow{2}{*}{\textit{\begin{tabular}[c]{@{}c@{}}Liveness \\ Acc.{[}\%{]}\end{tabular}}} & \multicolumn{1}{c|}{\multirow{2}{*}{\textit{\begin{tabular}[c]{@{}c@{}}Ferrlive\\ {[}\%{]}\end{tabular}}}} & \multicolumn{1}{c|}{\multirow{2}{*}{\textit{\begin{tabular}[c]{@{}c@{}}Ferrfake\\ {[}\%{]}\end{tabular}}}} & \multirow{2}{*}{\textit{\begin{tabular}[c]{@{}c@{}}Liveness\\ Acc.{[}\%{]}\end{tabular}}} & \multicolumn{1}{c|}{\multirow{2}{*}{\textit{\begin{tabular}[c]{@{}c@{}}Ferrlive\\ {[}\%{]}\end{tabular}}}} & \multicolumn{1}{c|}{\multirow{2}{*}{\textit{\begin{tabular}[c]{@{}c@{}}Ferrfake\\ {[}\%{]}\end{tabular}}}} & \multirow{2}{*}{\textit{\begin{tabular}[c]{@{}c@{}}Liveness \\ Acc.{[}\%{]}\end{tabular}}} & \multicolumn{1}{c|}{}\\
\multicolumn{1}{|c||}{}    & \multicolumn{1}{c|}{} & \multicolumn{1}{c|}{} &        & \multicolumn{1}{c|}{} & \multicolumn{1}{c|}{} &       & \multicolumn{1}{c|}{} & \multicolumn{1}{c|}{} &        & \multicolumn{1}{c|}{}\\ \hline
LivDet19\_CNN & 0.88    & 0.41    & 99.38& 8.54    & 40.59   & 73.98 & 12.22   & 0.37    & 93.98& 89.11    \\
unina         & 8.04    & 6.78    & 92.65& 3.53    & 81.85   & 53.77 & 0.51    & 5.61    & 96.82& 81.08    \\
PADUnkFv      & 3.24    & 1.55    & 97.68& 4.80    & 7.67    & 93.63 & 3.64    & 2.02    & 97.21& 96.17    \\
FSB  & 0.49    & 0.08    & 99.73& 7.37    & 23.84   & 83.64 & 4.95    & 0.28    & 97.50& 93.62    \\
JLW\_LivDet   & 0.39    & 1.14    & 99.20& 7.75    & 13.96   & 88.86 & 4.75    & 0.55    & 97.45& 95.17    \\
JLWa& 0.39    & 1.14    & 99.20& 7.75    & 14.06   & 88.81 & 4.75    & 0.55    & 97.45& 95.15    \\
JLWs         & 0.39    & 1.14    & 99.20& 7.75    & 14.06   & 88.81 & 4.75    & 0.55    & 97.45& 95.15    \\ 
BRLFptLivDet  & 11.08   & 0.33    & 94.79& 9.62    & 38.40   & 74.68 & 10.91   & 2.76    & 93.36& 87.61    \\
halekim      & 4.22    & 62.17   & 64.17& 8.24    & 15.70   & 87.70 & 28.79   & 28.03   & 71.61& 74.49    \\
JungCNN       & 1.08    & 0.82    & 99.06& 1.67    & 33.03   & 81.23 & 1.11    & 0.64    & 99.13& 93.14    \\
ZJUT\_Det\_A  & 0.39    & 1.14    & 99.20& 7.75    & 14.15   & 88.77 & 4.65    & 0.55    & 97.50& 95.16    \\
ZJUT\_Det\_S  & 0.39    & 1.14    & 99.20& 7.75    & 14.15   & 88.77 & 4.65    & 0.55    & 97.50& 95.16    \\
 \hline
\end{tabular}
}
\caption{Liveness accuracy of the algorithms on the test sets. For each dataset  the rate of misclassified live  and fake fingerprints are reported. The last column is relative to the average of the total accuracy on the three datasets.}
\label{tab:overalaccuracy}
\end{table*}
The Table \ref{tab:overall} shows the results of the integrated systems, in particular the accuracy for each dataset related to correctly recognized impostors and genuine and the total accuracy calculated as an average between the three datasets.
It can be seen that generally on the Green Bit scanner the IMG\_accuracy is higher than IMI\_accuracy, and this trend changes completely in the other two sensors. This can be related to the acquisition area of the sensors used: in particular, the Green Bit device has an area that covers the entire surface of the finger. A larger area allows to extract more minutiae and therefore have more points to compare.
In particular, the Green Bit sensor area is almost double compared to the other sensors (Table  \ref{tab:sensors}).

Regarding the ZJUT DET algorithm, of which two versions that differed only on the parameters setting have been presented, it is worth noting that these two versions return very different accuracies.
From cross-analysis of liveness and IMS results \ref{tab:overalaccuracy}-\ref{tab:overall}, it is interesting to highlight how the performance of some integrated systems is strongly influenced by those of liveness, such as halekim, while other algorithms, despite having problems distinguishing false from true, are able to return a correct output IMS (see DP LivDet19\_CNN results). Others, as ZJUT\_Det\_S, despite having high liveness results, lose accuracy with the matching module. This can depend on various factors, but the fusion rule of the matching and liveness results is very important.
For those algorithms whose performance in Presentation Attacks Detection appears as good, it could be the case to design a better integration algorithm in order to avoid the most significant problem in fingerprint liveness detection: the user misclassification due to errors introduced by the PAD analysis suggests that many efforts are still necessary.

\begin{table*}[]
\centering
\resizebox{\textwidth}{!}{%
\begin{tabular}{|l||c|c|c||c|c|c||c|c|c||c|}
\hline
\multirow{2}{*}{\textbf{Algorithm}} & \multicolumn{3}{c||}{\textbf{Green Bit}}      & \multicolumn{3}{c||}{\textbf{Digital Persona}} & \multicolumn{3}{c||}{\textbf{Orcanthus}}      & \multirow{2}{*}{\textbf{Overall}} \\ \cline{2-10}
                                    & \textit{IMG\_acc} & \textit{IMI\_acc} & \textit{Total\_acc} & \textit{IMG\_acc}  & \textit{IMI\_acc} & \textit{Total\_acc} & \textit{IMG\_acc} & \textit{IMI\_acc} & \textit{Total\_acc} &                                    \\ \hline  \hline
LivDet19\_CNN                       & 82.16        & 99.96        & 95.67          & 67.34         & 98.30        & 90.84          & 67.35        & 100.00       & 91.78          & 92.77                              \\ 
unina                               & 88.36        & 97.81        & 95.54          & 71.37         & 92.78        & 87.62          & 76.69        & 99.58        & 93.82          & 92.32                              \\ 
JLWa                              & 99.12        & 96.60        & 97.21          & 89.69         & 90.31        & 90.16          & 92.55        & 95.97        & 95.11          & 94.16                              \\ 
JLWs                                & 99.24        & 98.72        & 98.85          & 88.48         & 95.75        & 94.00          & 91.84        & 99.80        & 97.80          & 96.88                              \\ 
halekim                            & 93.43        & 82.43        & 85.08          & 82.32         & 96.84        & 93.35          & 61.72        & 96.57        & 87.80          & 88.74                              \\ 
JungCNN                             & 98.46        & 98.26        & 98.31          & 96.66         & 85.98        & 88.56          & 96.97        & 98.33        & 97.99          & 94.95                              \\ 
ZJUT\_Det\_A                        & 99.51        & 95.89        & 96.76          & 93.07         & 86.65        & 89.37          & 92.55        & 96.01        & 95.14          & 93.76                              \\ 
ZJUT\_Det\_S                        & 99.66        & 30.24        & 46.97          & 92.76         & 58.43        & 66.70          & 96.67        & 64.50        & 72.60          & 62.09                              \\ 
\hline
\end{tabular}
}
\caption{IMS accuracy of the algorithms on the test sets [\%]. For each dataset  the rate of correctly classified  genuine live fingerprints (IMG) and the rate  of  correctly  classified  impostor live or genuine fake fingerprints (IMI) are reported. The last column is relative to the average of the total accuracy on the three datasets.}
\label{tab:overall}
\end{table*}

\section{Conclusion}
    
\label{sec:Conclusion}
The sixth edition of LivDet marks a transition point compared to previous editions. As a matter of fact, this edition was intended to investigate how the integration of a liveness detection system with a fingerprint authentication system might impact on the performance.

The results show that many of the integrated systems presented achieved a performance of over 90\%. We hypothesise that the low performance were due to a low liveness performance or incorrect setting of the matcher parameters.

To sum up, this six edition is the first one that investigates a complete verification system integrated with a PAD, based on the state-of-the-art approaches in such topic.
We hope to have added some interesting insights for all researchers and companies working in this challenging matter. From our side, we want to continue on this path, by taking especially care, in the next edition, on the spoofs quality and materials explicitly synthesised to be ``spoofing-effective''.

\section{Acknowledgements}

This edition has been possible thanks to the sponsorship of the Green Bit company. 
We want to thank Stephanie Schuckers, who suggested us the initial insights to the fabrication of fake fingerprints in 2006 and cooperated to the organization of the first competition (2009) up to the following ones (2011, 2013, 2015). 
Her help was significant for allowing us to autonomously gain, over time, a deep knowledge on the problem of fingerprint presentation attacks detection and contributed to the state-of-the-art.

\end{document}